%% file: icdm.tex
\documentclass[10pt, conference, compsocconf]{ieeetran}
\usepackage{times}
\usepackage{helvet}
\usepackage{courier}
\usepackage[english]{babel}
\usepackage[utf8x]{inputenc}
\usepackage{amsmath}
\usepackage[equations=false]{resizegather}

\usepackage{amssymb}
\usepackage{graphicx}
\usepackage{color}
\usepackage[colorinlistoftodos]{todonotes}

\usepackage{hyperref}
\usepackage{verbatim}
\usepackage{enumitem}
\usepackage{array}
\usepackage{euscript}
\usepackage{xspace}

\usepackage{multirow}
\usepackage{algorithm}
\usepackage[noend]{algpseudocode}
\usepackage{caption}

\newcommand{\bs}{\boldsymbol}
\newcommand{\mc}{\mathcal}
\newcommand{\mf}{\mathfrak}
\newcommand{\NTT}{N}
\newcommand{\NTW}{\mathfrak{T}}
\newcommand{\NUT}{\mathfrak{U}}

\newcommand{\Words}{{\cal W}}
\newcommand{\HMHP}{\textsf{HMHP}\xspace}
\newcommand{\HTM}{\textsf{HTM}\xspace}
\newcommand{\HTL}{{\textsf{HWK+Diag}}\xspace}
\newcommand{\NWL}{{\textsf{HWK$\times$LDA}}\xspace}
\newcommand{\real}{\texttt{Twitter}\xspace}
\newcommand{\semi}{\texttt{SemiSynth}\xspace}
 
\begin{document}
%
\title{Discovering Topical Interactions in Text-based Cascades using Hidden Markov Hawkes Processes}

\author{
	\IEEEauthorblockN{Jayesh Choudhari, Anirban Dasgupta}
	\IEEEauthorblockA{IIT Gandhinagar, India\\
		Email: \{choudhari.jayesh, anirbandg\}@iitgn.ac.in}
	\and
	\IEEEauthorblockN{Indrajit Bhattacharya}
	\IEEEauthorblockA{TCS Research, India\\
		Email: b.indrajit@tcs.com}
	\and
	\IEEEauthorblockN{Srikanta Bedathur}
	\IEEEauthorblockA{IIT Delhi, India\\
		Email: srikanta@cse.iitd.ac.in}
}

\maketitle
\input{abstract}

\section{Introduction}
\input{intro}

\section{Model}
\input{model}

\section{Inference}
\label{sec:inf}
\input{inference}

\section{Experiments}
\input{expt}

\section{Related Work}
\input{rw}

\section{Conclusion}
\input{conclusion}

\bibliographystyle{plain}
\bibliography{icdm}

\end{document}

%% file: abstract.tex
\begin{abstract}
Social media conversations unfold based on complex interactions between users, topics and time.
While recent models have been proposed to capture network strengths between users, users' topical preferences and temporal patterns between posting and response times, interaction patterns between topics has not been studied.
We argue that social media conversations naturally involve interacting rather than independent topics.
Modeling such topical interaction patterns can additionally help in inference of latent variables in the data such as diffusion parents and topics of events.
We propose the Hidden Markov Hawkes Process (\HMHP) that incorporates topical Markov Chains within Hawkes processes to jointly model topical interactions along with user-user and user-topic patterns.
We propose a Gibbs sampling algorithm for \HMHP that jointly infers the network strengths, diffusion paths, the topics of the posts as well as the topic-topic interactions.
We show using experiments on real and semi-synthetic data that \HMHP is able to generalize better and recover the network strengths, topics and diffusion paths more accurately that state-of-the-art baselines.
More interestingly, \HMHP finds insightful interactions between topics in real tweets which no existing model is able to do. 
This can potentially lead to actionable insights enabling, e.g., user targeting for influence maximization. 
\end{abstract}

%% file: intro.tex
A popular area of recent research has been the study of information diffusion cascades, where information spreads over a social network when a `parent' event from one infected node influences a `child' event at neighboring node \cite{du:kdd15,he:icml15,zhao:kdd2015,gao:wsdm2015,guille2013information}.
The action of propagating information between two neighboring nodes depends on various factors, such as the strength of influence between the nodes, the topical content of the parent event and the extent of interest of the child node towards that topic. 
Explosion of social media data has made it possible to analyze and evaluate different models that seek to explain such information cascades.
However, many relevant variables such as the network influence strengths, the identity of influencing or parent event for any event, and the actual topics are typically unobserved for most social network data.
Therefore, these need to be inferred before any analysis of the diffusion patterns can be performed.

A recent body of work in this area has proposed increasingly sophisticated models for information cascades.
To account for the temporal burstiness of events, Hawkes processes, which are self-exciting point processes, have been proposed to model their time stamps \cite{rizoiu:arxiv2017}.
Also, influences travel more quickly over stronger social ties.
This is modeled by the Network Hawkes process \cite{linderman:icml14} by incorporating connection strengths between users into the intensity function of the Hawkes process. 
When events additionally have associated textual data, parent and child events in a cascade are typically similar in their topical content.
The Hawkes Topic Model \cite{he:icml15} captures this by combining the Network Hawkes process with a LDA-like generative process for the textual content where the topic mixture for a child event is close to that for a parent event.
Given partially observed data, inference algorithms for both the Network Hawkes and the Hawkes Topic models recover latent parent identities for events and the network strengths.
In addition, the Hawkes Topic model enables recovery of the latent topics of the events.
The main strength of this inference algorithm is its collective nature - the recovery of topics, parents and network strengths reinforce each other.

Despite its strengths, the Hawkes Topic Model fails to capture one important aspect of the richness of information cascades.
Typically, there are sequential patterns in the textual content of different events within a cascade. 
In terms of topical representation of the events, topics of parent and child events display repeating patterns. 
Consider the following pair of parent-child tweets detected by our model: 
\textsc{``Though seemingly delicate, monarch \#butterflies are remarkably resilient \& their decline can be reversed \#pesticides''} and \textsc{``EPA Set to Reveal Tough New Sulfur Emissions Rule  \#climatechange \#sustainability \#pollution''}.
It is not obvious how to place these two tweets in the same topic. 
Our model places them in different topics (say {\em pesticides and butterflies} and {\em climate change and pollution}) and detects that this topic pair appears frequently in other parent-child tweets in our data.
Adding to this their posting time information, it has sufficient evidence to detect a parent-child relation between this pair. 

However, just like topic distributions, topic interaction patterns are also typically latent, and need to be inferred.
Interestingly, inferring topic interaction patterns in turn benefits from more accurate parent and topic assignment to events.
This calls for joint inference of topic interaction patterns and the other latent variables.

We propose a generative model for textual cascades that captures topical interactions in addition to temporal burstiness, network strengths, and user topic preferences.
Temporal and network burstiness is captured using a Hawkes Process over the social network \cite{linderman:icml14}.
Topical interactions are modeled by combing this process with a Markov Chain over event topics along a path of the diffusion cascade.
Such topical Markov Chains have appeared in the literature \cite{griffiths:nips05,gruber:aistats07,barbieri2017survival,barbieri13prob,Du2012}, but in the very different context of modeling word sequences in a document.
Our model effectively integrates topical Markov Chains with the Network Hawkes process and we call it the Hidden Markov Hawkes Process (\HMHP).

Making use of conjugate priors, we derive a simple and efficient collapsed Gibbs Sampling algorithm for inference using our model.
This algorithm is significantly more scalable than the variational algorithm for the Hawkes Topic Model, and allows us to analyze large collections of real tweets.
We validate using experiments that \HMHP fits information cascades better compared to models that do not consider topical interactions.
With the aid of semi-synthetic data, we show that modeling topical interactions indeed leads to more accurate identification of event parents and topics.
More importantly, \HMHP is able to identify interesting topical interactions in real tweets, which is beyond the capability of any existing information diffusion model. 

One of the underlying aims of modeling information cascades is to develop accurate models that can predict the growth and virality of cascades on specific topics started by specific users-- such models can then be used in practice to select a set of users to incentivize to start a conversation about the topics desired. Our modeling of the topic-topic interaction can broaden this approach by providing access to users who generate content about related set of topics, from which, ultimately, the conversations tend to drift to the topics of interest. Mechanisms that can estimate such possible related candidate topics to target could thus be practically very useful, in addition to providing unique insights about how conversations evolve on social networks.


%% file: model.tex
\label{sec:model}
We consider a set of nodes $V = \{1, \ldots, n\}$, representing content producers, and a set of directed edges $E$ among them, 
representing the underlying graph using which information propagates. 
For every edge $(u,v)$, $W_{uv}$ denotes the weight of the edge, capturing the extent of influence between content producers or users $u$ and $v$.
While we assume the underlying graph structure
to be observed, the actual weights $W_{uv}$ are not. This captures the intuition that while the graph structure may be known, 
the actual measure of influence that one user has on another is not. 
For each event $e$, representing e.g. a social media post, we observe its posting time 
$t_e$, the user $c_e$ who creates the post and the document $d_e$ that is associated with the post. 
The posting time follows a Hawkes Process \cite{simma2010} incorporating user-user edge weights \cite{linderman:icml14}.
Event $e$ could be {\em spontaneous}, or a {\em diffusion} event, meaning that it is triggered by some recent event, which we call its parent $z_e$, created by one of the followees of user $c_e$.

The document $d_e$ for $e$ is drawn using a topic-model over a  vocabulary of size $\Words$. 
For spontaneous events, the topic choice $\eta_e$ depends on the topic preference of the user $c_e$.
We deviate from existing literature in modeling the topic of a diffusion event.
Instead of being identical or `close' to the topic of the triggering event \cite{he:icml15}, the diffusion event may be on a `related' topic.
We model this transition between related topics using a Markov Chain over topics, involving a topic transition matrix ${\mc{T}}$. 
This enables us to capture repeating patterns in topical transitions between parent and child events.

The generated event $e$ is then consumed by each of the followers of the user $c_e$, 
thereby triggering them in turn to create multiple events and producing an information cascade. 
Since the topical sequence is `hidden' and observed only indirectly through the content of the events, 
we call our model the Hidden Markov Hawkes Process (HMHP).
We next describe the details of our generative model.
This can be broken up into two distinct stages - generating the cascade events, and generating the event documents.


\subsection{Generating Cascade Events}
We define an event $e$ by a tuple
$(t_e, c_e, z_e, d_e)$ where $t_e$ indicates the time for event $e$, $c_e$ the id of the creator node and $z_e$ the unique parent 
event that triggered the creation of event $e$, set to $0$ if the event $e$ is spontaneous, and $d_e$ is the textual document. 
The generative model for \HMHP consists of two phases. 
We first generate $(t_e, c_e, z_e)$ for all events using a multivariate Hawkes process (MHP) following existing models \cite{he:icml15,linderman:icml14}, and then use a hidden Markov process 
based topic model to generate all the associated documents $d_e$. 
A multivariate Hawkes process  
models the fact that the users can mutually excite each other to produce events. 
A Hawkes process can also be represented
as a superposition of Poisson processes~\cite{simma2010}. 
For each node $v$, we define $\lambda_v(t)$, a rate at time $t$,
as a superposition of the base intensity $\mu_v$ for user $v$, and the impulse responses 
for each event $e_n$ that has happened at time $t_n$ at a followee node $c_n$.
\begin{align}
\label{eq:hawkes_v}
\lambda_{v}(t) = \mu_{v}(t) + \displaystyle \sum_{n=1}^{|\mc{H}_{t^-}|} h_{c_n, v}(t-t_n)
\end{align}
where $h_{c_n, v}(t-t_n)$ is the impulse response of user $c_n$ on the user $v$, and $\mc{H}_{t^-}$ 
is the history of events upto time $t$. The impulse response can be decomposed 
as the product of the influence $W_{uv}$ of user $u$ on $v$, and a time-kernel as follows:
\begin{align}
\label{eq:impulse_res}
h_{u, v}(\Delta t) = W_{u,v} f(\Delta t)
\end{align}
We note that while there has been a number of recent works on modeling the time-kernel
effectively~\cite{linderman:icml14,du:kdd15,he:icml15}, we use a simple exponential kernel $f(\Delta t ) = exp(-\Delta t)$, as this is not 
the main thrust of our work. Following~\cite{simma2010}, we generate the events using a level-wise generation process.
Level $0$, denoted as $\Pi_0$, contains all the spontaneous events, generated using the base rates of the users. 
The events $\Pi_l$ at level $l$, are generated according to the following non-homogenous Poisson process
\begin{align}
\label{eq:generate_events_l}
\Pi_{l} \sim Poisson \left(\displaystyle \sum_{(t_n, c_n, z_n) \in \Pi_{l-1}} h_{c_n,\cdot}(t-t_n)\right)
\end{align}
The above process can be simulated by the following algorithm-- for each event $e_n = (t_n, c_n, z_n) \in \Pi_{l-1}$,
and for each neighbor $v$ of $c_n$, we draw timestamps according to the non-homogeneous Poisson process-- $Poisson(h_{c_n, v}(t-t_n))$;
we generate an event $e$ for each of these timestamps, and set $z_{e} = e_n$ (parent) and $c_e = v$ (producer node). 

\subsection{Generating Event Documents}
The main focus of our model is to capture repeating patterns in topical transitions between parent and child events.
We posit the existence of a fixed number of topics $K$.
The topics, denoted $\{\bs{\zeta}_k\}$, are assumed to be probability distributions over words (vocabulary with size $\mc{W}$)
and are generated from a Dirichlet distribution. 
Since our data of interest is tweets (short documents), we model a document at event $e$ as having a single hidden topic $\eta_e$. 
We also assume the existence of a topic-topic interaction matrix $\bs{\mc{T}}$, where $\bs{\mc{T}}_{k}$, again sampled from a Dirichlet distribution, denotes the distribution over `child topics' for a `parent topic' $k$.
In order to generate the document $d_e$ and event $e$, we first 
follow a Markov process for sampling the topic $\eta_e$ of the document conditioned on the topic of its parent event $\eta_{z_e}$,
followed by sampling the words according to the chosen topic. 
For spontaneous events occurring at any user node $u$, the topic for its document is sampled randomly from the preferred distribution over topics $\bs{\phi}_{u}$ for node $u$. 

An important consideration in the design of our model is the use of conjugate priors.
As we will see in the Inference section such priors play a crucial role in the design of efficient and simple sampling-based inference algorithms.
Models such as HTM \cite{he:icml15}, which have to sacrifice conjugacy to model data complexity, have to resort to more complex variational inference strategies.

\begin{figure}[]
  \centering
  \captionsetup{justification=centering}
  \includegraphics[width=\linewidth]{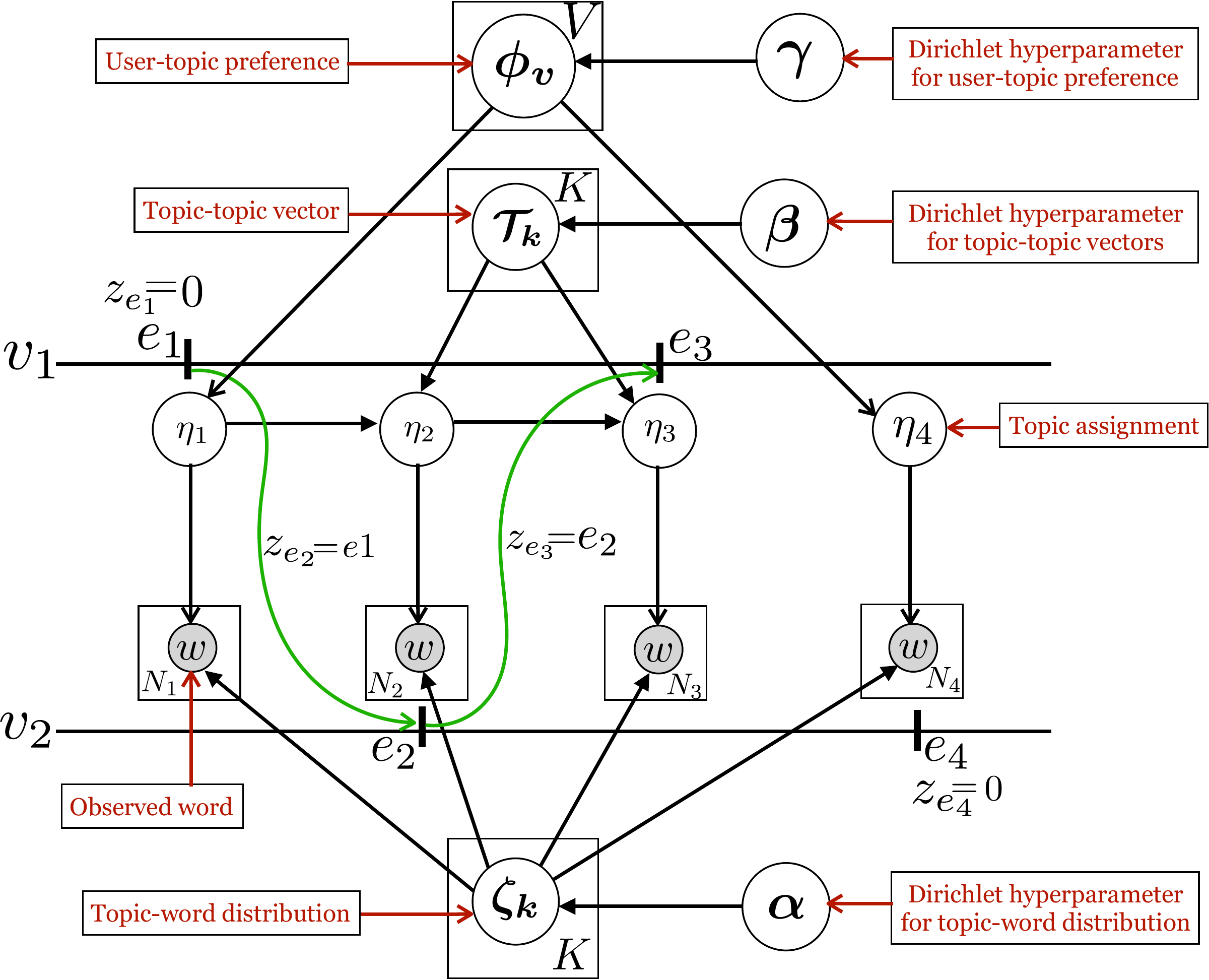}
  \caption{Graphical Model for \HMHP (shown for 2 nodes and only 4 sample events)}
  \label{fig:graphical_model}
\end{figure}

Figure~\ref{fig:graphical_model} shows the plate diagram of the generative model.
We summarize the entire generative process below.
\begin{enumerate}
\item Generate $(t_e, c_e, z_e)$ for all events according to the process described in previous sub-section.
\item For each topic $k$: sample $\bs{\zeta}_k \sim Dir_{\Words}(\bs{\alpha})$ 
\item For each topic $k$: sample $\bs{\mc{T}}_{k} \sim Dir_{K}(\bs{\beta})$ 
\item For each node $v$: sample $\bs{\phi}_{v} \sim Dir_{K}(\bs{\gamma})$ 
\item For each event $e$ at node $c_e=v$:
\begin{enumerate}
	\item
	\begin{enumerate}
	\item \textbf{if} $z_e=0$ (level $0$ event):\\
	draw a topic $\eta_{e} \sim Discrete_{K}(\bs{\phi}_{v})$
	\item \textbf{else}: \\
	draw a topic $\eta_{e} \sim Discrete_{K}(\bs{\mc{T}}_{\eta_{z_e}})$
	\end{enumerate}
	\item Sample document length $\NTT_e \sim Poisson(\lambda)$
	\item For $w = 1 \dots \NTT_e$: draw word $x_{e,w} \sim Discrete_{\Words}(\bs{\zeta}_{\eta_{e}})$
	\end{enumerate}
\end{enumerate}
The resultant joint likelihood can we written as follows:
		\begin{gather}
				P(E, \bs{\Phi}, \bs{\mc{T}}, \bs{\zeta}, \bs{\eta}, \bs{z} \mid \bs{\alpha}, \bs{\beta}, \bs{\gamma}, \bs{W}, \bs{\mu}) = \nonumber \\ 
				\prod_{v\in V} P(\bs{\phi}_{v} \mid \bs{\gamma}) \times  \prod_{k=1}^{K} P(\bs{\zeta}_{k} \mid \bs{\alpha}) \times \prod_{k=1}^{K}  P(\bs{\mc{T}}_{k} \mid \bs{\beta}) \nonumber \\
				\times \prod_{e \in E} \left\{ \left[ \prod_{e':t_{e'}<t_e} P(\eta_e|\bs{\mc{T}}_{\eta_{z_e}})^{\delta_{z_{e}, e'}} \right] P(\eta_e|\bs{\phi}_{v})^{\delta_{z_e,0}} \right\}  \nonumber\\
					\times \prod_{e\in E}\left[ \prod_{w=1}^{\NTT_e} P(x_{e,w}|\eta_e, \bs{\zeta}_{\eta_e}) \right] \nonumber \\
					\times \prod_{v \in V} \left[ exp \left( -\int_{0}^{T} \mu_{v}(\tau) d\tau \right) \prod_{e \in E} \mu_{v}(t_e)^{\delta_{c_e,v} \delta_{z_e,0} } \right] \nonumber \\
		\label{eq:cond_probability}
                  \times \prod_{e\in E} \prod_{v \in V} \left[ exp \left( -\int_{t_e}^{T} h_{c_e, v}(\Delta\tau) d\tau \right) \prod_{{e'} \in E}  h_{c_e, c_{e'}}(\Delta(t_{e'}))^{\delta_{c_{e'},v} \delta_{z_{e'}, e} }  \right] 
		\end{gather}

Here, $\delta_{c_{e},v}$ is an indicator for the event $e$ being triggered at node $v$, $\delta_{z_{e'}, e}$ is an indicator for event $e$ being parent of event $e'$, $\Delta\tau$ is $(\tau - t_e)$, $\Delta(t_{e'})$ is $(t_{e'} - t_e)$ and $T$ is the time horizon.
The first line correspond to drawing user topic preference vectors, vectors for word distributions over topics, and topic-topic probability vectors from the corresponding Dirichlet. 
The second line corresponds to the probability of selecting the event topic depending on the topic of the parent event. 
The third line is for generating the event words given the event topic and topic distribution. 
The fourth line corresponds to the base intensity of the Hawkes processes and the fifth line captures the impulse response. The last term can be interpreted as the impulse response of each event $e$ on all events $e'$ triggered at node $v$.

%% file: inference.tex
Given model definition, the underlying graph structure and the observable features of the events $E$, our task is to identify the latent variables associated with each event and also estimate the hidden parameters of the model.
The latent event variables are the topic $\eta_e$ and the diffusion parent $z_e$ for identifying the cascade diffusion structure. 
For each event $e$, we will either decide the event to be spontaneous ($z_e=0$), or identify the unique {\em parent event} $e'$ that triggered the creation of $e$ ($z_e=e'$).
The process parameters to be estimated include the user-user influence values $W_{uv}$, the topic transition matrix ${\mc{T}}$, and the user-topic preferences.

Since exact inference is intractable, we perform inference using Gibbs Sampling, where the strategy is to iteratively sample the value of each hidden variable from its conditional distribution given the current values of all the other variables, and continue this process until convergence.
This process can be made more simple and efficient for our model.
By making use of the conjugate priors, we can perform collapsed Gibbs Sampling, where we integrate out some of the continuous valued parameter variables from the likelihood function in Eqn.\ref{eq:cond_probability}.
Specifically, we integrate out the topic distributions over words $\bs{\zeta}$, the topic interaction distributions $\bs{\mc{T}}$ and the user-topic preference distributions $\bs{\Phi}$.
The continuous valued latent variables that remain are the connection strengths ${W}_{u,v}$.
The resulting collapsed Gibbs sampling algorithm iteratively samples individual topic and parent assignments and the connection strengths from their conditional distributions given current assignments to all other variables until convergence.
The parameter variables that were integrated out are estimated from the samples of the remaining variables upon convergence.
We next describe the conditional distributions for sampling the different variables in the algorithm.
The overall algorithm is described in Algo.\ref{alg:gibbsSampler}.

\paragraph{Topic assignment} The conditional probability for topic $\eta_{e}$ of a diffusion event $e$ ($z_e \neq 0$) being $k$ when the current parent has topic $k'$ is the following:

\begin{equation}								\label{eq:topic_assignment_prob_with_parent_event}
\begin{array}{*1{>{\displaystyle}l}}
P(\eta_e = k \mid \{x_{e\cdot}\}, \eta_{z_e} = k', 		\bs{\eta}_{\setminus z_e}, \{z_e\}) \propto \\
\dfrac { \beta_{k} + \NTT_{k',k}^{(\lnot (z_e, e))} } { (\sum_{l} \beta_l) + \bs{\NTT}_{k'}^{(\lnot (z_e,e))}} \times 
\dfrac { \prod_{l' = 1}^K \prod_{i = 0}^{ \NTT_{k,l'}^{(C_{e})} - 1}( \beta_{l'} +  \NTT_{k,l'}^{(\lnot C_e)} + i) } { \prod_{i = 0}^{ \bs{\NTT}_{k}^{(C_{e})} - 1 } ( (\sum_{l'} \beta_{l'}) +  \bs{\NTT}_{k}^{\lnot C_e} + i ) } \\~\\
\qquad \qquad \times  
\dfrac{\prod_{w \in d_e} \prod_{i = 0}^{N_{e}^{w}-1} (\alpha_w + \NTW_{k, w}^{\lnot e} + i)}
{\prod_{i=0}^{N_{e} - 1} ((\sum_{w \in \Words} \alpha_w) + \NTW_{k}^{\lnot e} + i) }
\end{array}
\end{equation}
Here $\NTT_{k',k}^{(C)}$ denotes the number of parent-child event pairs with topics $k$ and $k'$, $(\lnot (z_e, e))$ denotes all edges excluding $(z_e,e)$, $C_e$ is the set of edges from event $e$ to its child events ($\lnot C_e$ being its complement), and $\NTW_{k, w}^{\lnot e}$ is the number of occurrences of word $w$ under topic $k$ in events other than $e$. Further, $\bs{N}_k^C=\sum_{k'} N_{k,k'}^{C}$, and $N^w_{e}$ is count of $w$ in $d_e$. 
The first term is the conditional probability of transitioning from parent topic $k'$ to this event's topic $k$, the second term is that of transitioning from this topic $k$ to each child event's topic $l'$, and the third term is that of observing the words in the event document given topic $k$.
It is important to observe how this conditional distribution pools together three different sources of evidence for an event's topic.
Even when the document words do not provide sufficient evidence for the topic, the parent and children topics taken together can significantly reduce the uncertainty.

For spontaneous events ($z_e = 0$), the conditional probability looks very similar to Eqn. \ref{eq:topic_assignment_prob_with_parent_event}.
Only the first term changes to $\dfrac{\gamma_{k} + \NUT_{v,k}^{(\lnot e)} }{(\sum_{k} \gamma_k) + \bs{\NUT}_{v}^{(\lnot e)}}$, where $\NUT_{v,k}^{(\lnot e)}$ is the number of events by user $v$ with topic $k$ discounting event $e$. 
This term captures the probability of node $v$ picking topic $k$ from among its preferred topics.


\paragraph{Parent assignment} The conditional probability of event event $e'$ being the parent $z_e$ of an event $e$ looks as follows:
				
\begin{equation*}
\begin{array}{*1{>{\displaystyle}l}}
P(z_e = e' | E_{t}, \bs{z}_{\lnot e}, \bs{{W}}, \bs{\mu}) \\
	\quad \quad \propto \dfrac{(\beta_k + \NTT_{k',k} - 1)  }{((\sum_{k=1}^{K}\beta_k) + \bs{\NTT}_{k'} - 1)} \times h_{u_{e'}, u_{e}}(t_{e} - t_{e'})
\end{array}
\end{equation*} 
Here the first term is the transition probability from topic $k'$ of the proposed parent event $e'$ to this event's topic $k$.
The second term is the probability of $t_e$ being the time of this event $e$ given the occurrence time $t_{e'}$ of the proposed parent event $e'$.
As with topic identification, we see that evidence for the parent now comes from two different sources.
When there is uncertainty about the parent based on the event time, common patterns of topic transitions between existing parent-child events helps in identifying the right parent.

The conditional probability of event $e$ being a spontaneous event with no parent is given as:
\begin{equation*}
\begin{array}{*1{>{\displaystyle}l}}
P(z_e = 0 | E_{t}, \bs{z}_{\lnot e}, \bs{{W}}, \bs{\mu}) 
	\propto \\
    \quad \quad \dfrac{(\gamma_k + \NUT_{u_e,k} - 1)  }{((\sum_{k=1}^{K}\gamma_k) + 	\bs{\NUT}_{u_e} - 1)} \times \mu_{u_e}(t_{e})
\end{array}
\end{equation*}
Here again, we have two terms related to the topic and the event time.
But the topic term captures the probability of user $u_e$ spontaneously picking topic $k$ for an event, and the time term captures the probability of the same user spontaneously generating an event at time $t_e$ given its base intensity.

In theory, every preceding event is a candidate parent for an event $e$.
We limit the number of possible parent candidates for an event by the time interval between events (1 day) to a maximum of 100 candidates.


\paragraph{Updating network strengths} For the network strength ${W}_{u,v}$, using a Gamma prior $Gamma(\alpha, \beta)$, the posterior distribution can be approximated as follows:
			\begin{equation*}
				\begin{array}{*2{>{\displaystyle}l}{p{3cm}}}
					P({W}_{u,v} = x \mid E_{t}^{(u,v)}, \bs{z}) \propto 
                    & x^{\alpha_{1}} \exp (-x\beta_{1})
				\end{array} 	
			\end{equation*}
where $\alpha_{1} = (\NTT_{u,v} + \alpha - 1)$ and $\beta_1 = (\NTT_u + \frac{1}{\beta})^{-1}$, and $\NTT_{u,v}$ is the number of parent-child events pairs between nodes $u$ and $v$, and $\NTT_u$ is the number of events at node $u$.
This is again a Gamma distribution $Gamma(\alpha_{1}, \beta_{1})$.
Instead of sampling $W_{uv}$, we set it to be the mean of the corresponding Gamma distribution.
Note that in each iteration, we update $W_{uv}$ only for edges that have at least one parent-child influence.

A practical issue with estimating $W_{uv}$ is that in real datasets, most edges have very few (typically just one) influence propagation event.
This makes statistical estimation of their strengths infeasible.
To get around this problem, we share parameters across edges.
While there may be many ways to group `similar' edges, 
we group together edges that have the same value for the tuple (out-degree(source), in-degree(destination)). 
The intuition is that the influence of the edge $(u,v)$ is determined uniquely by the the popularity (outdegree) of $u$ and the number of different influencers (indegree) of $v$. 
We then pool data from all edges in a group and estimate a single connection strength for a group. 

Once the Markov chain has converged, the parameters that were integrated out are estimated using the samples:  
					$$\hat{\zeta}_{k,r} = \frac{\NTW_{k,r} + \alpha_r}{\sum_{r' = 1}^{|\mf{W}|}\NTW_{k,r'} + \alpha_{r'}} \quad \quad \hat{\phi}_{v,k} = \frac{\NUT_{v,k} + \gamma_k}{\sum_{k' = 1}^{K} \NUT_{v,k'} + \gamma_{k'}}$$ 
					$$\hat{\mc{T}}_{k,k'} = \frac{\NTT_{k,k'} + \beta_{k'}}{\sum_{t = 1}^{K} \NTT_{k,t} + \beta_{t}}$$
Here, $\NTW_{k,r}$ is the number of occurrences of word $r$ in events with topic $k$, $\NUT_{v,k}$ is the number of events posted by user $v$ with topic $k$, and $\NTT_{k,k'}$ is the number of parent-child events with topics $k$ and $k'$ respectively.

\begin{algorithm}
\caption{Gibbs Sampler}\label{alg:gibbsSampler}
\begin{algorithmic}[H]
\State Initialize $\eta_e$ for all events
\State Initialize $z_e$ for all events
\State Initialize $W_{u,v}$ for all $u,v$ in the followers map
\For{$iter=0$; $iter~!=maxIter$; $iter++$}  

\For{$e \in allEvents$} \Comment{Sampling topic}
\State $\eta_e \sim P(\eta_e = k \mid \bs{\eta}_{\lnot e}, \bs{z}, \{X\}, \bs{W}, \bs{\alpha}, \bs{\beta}, \bs{\gamma},\bs{\mu})$
\EndFor
    
\For{$e \in allEvents$} \Comment{Sampling Parent}
\State $z_e \sim P(z_e = e' | E_{t}, \bs{z}_{\lnot e}, \bs{\eta}, \bs{\alpha}, \bs{\beta}, \bs{\gamma}, \bs{W}, \bs{\mu})$
\EndFor    

\For{$(u,v) \in Edges$}
\Comment {Estimate user-user influence}
\State $W_{u,v} = Mean(Gamma(N_{uv} + \alpha', N_{u} + \beta'))$ 
\EndFor
\EndFor
\end{algorithmic}
\end{algorithm}

%% file: expt.tex
In this section, we empirically validate the strengths of \HMHP against competitive baselines over a large collection of real tweets as well as semi-synthetic data.
We first discuss the baseline algorithms for comparison, the datasets on which we evaluate these algorithms, the tasks and the evaluation measures, and finally the experimental results.
\subsection{Evaluated Models}
Recall that our model captures network structure, textual content and timestamp of the posts / tweets, and identifies the topics and parents of the posts, in addition to reconstructing the network connection strengths. 
Considering this, we evaluate and compare performance for the following models:
\begin{itemize}
 \item \HMHP: This is the full version of our model with Gibbs sampling-based inference. Recall that this performs all reconstruction tasks mentioned above jointly, while assuming a single topic for a post and topical interaction patterns.
\item \HTM: This is the state-of-the-art model ~\cite{he:icml15} closest to ours that addresses the same tasks.
The key modeling differences are two-fold: \HTM assumes a topical admixture for each document instead of a mixture, and secondly it models parent and child events to be `close' in terms of their topical admixture. 
As a result, it cannot capture any sequential pattern in the topical content of cascades. 
Additionally, it uses a reasonably complex variational inference strategy due to the absence of conjugacy in the prior distributions. 
We used an implementation of \HTM kindly provided to us by the authors~\cite{he:icml15}. 
\item \HTL: This is a simplification of our model where the topic-topic interaction is restricted to be diagonal. In other words, each topic interacts only with itself. However, the reconstruction tasks are still performed jointly.
This model helps in our evaluation in two different ways. 
First, it helps in assessing the importance of topical interactions.
Secondly, this serves as a crude approximation of \HTM with only one topic per document. 
Since the algorithm for the original \HTM did not scale for our larger datasets, we use this model as a surrogate for evaluation.
\item \NWL: This is motivated by the Network Hawkes model~\cite{linderman:icml14}, which jointly models the network strengths and the event timestamps, and therefore performs parent assignment and user-user influence estimation jointly. 
However, it does not model the textual content of the events.
Therefore, we augment it with an independent LDA mixture model~\cite{rigouste2006inference} (LDAMM) to model the textual content.  
While one view of this model is as an augmentation of the Network-Hawkes model, the other view is that of a simplification of our model, where the topic assignment is decoupled from parent assignment and network reconstruction.
Thus, comparison against this model helps in analyzing the importance of doing topic, parent and network strength estimation jointly. 
The network reconstruction component of \NWL is similar to the Network Hawkes model ~\cite{linderman:icml14}, with the only difference being that the hyper-parameters of the model are not estimated from the data. 
\end{itemize}

\subsection{Datasets} 
We perform experiments on two datasets, that we name \real and \semi. 
The \real dataset was created by crawling $7M$ seed nodes from Twitter for the months of March-April 2014. 
We restrict ourselves to $500K$ tweets corresponding to top 5K hashtags from the most prolific $1M$ users generated in a contiguous part of March 2014. 
For each tweet, we have the time stamp, creator-id and tweet-text. 
Note that gold-standard for the parameters or the event labels that we look to estimate is not available in this \real dataset. 
We do not know the true network connection strengths, event topics or cascade structure of the tweets.
While retweet information is available, it is important to point out that retweets form a very small fraction of the parent-child relations that we hope to infer.

Since we need gold-standard labels to evaluate performance of the models, we additionally create a \semi dataset using the generative process of our model while preserving statistics of the real data to the extent possible. 
From a sample of the \real data we retain the underlying set of nodes and the follower graph.
Then we `estimate' all the parameters of our model (the base rate per user, user-user influence matrix, the topic distributions, resulting topical interactions and the user-topic preferences) from \real data. 
The document lengths are randomly drawn from $Poisson(7)$, since $7$ was the average length of the tweets in the \real dataset.
We finally generate $5$ different samples of $1M$ events using our generative model. 
All empirical evaluations are averaged over the $5$ generated samples (together termed as \semi). 

The details of the parameter estimations are as follows. 
We assign as the parent for a tweet $e$ from user $u$ the `closest in time' preceding tweet from the followees of user $u$ for the last 1 day. 
If this set is empty, then $e$ is marked as a spontaneous event. 
Topics to every tweet are assigned by fitting a latent Dirichlet analysis based mixture model (LDAMM~\cite{rigouste2006inference}) with $100$ topics.
From the parent and topic assignments, we get the network strengths, user-topic preferences, topic-topic interactions and topic-word distributions.
For ${W}_{uv}$ estimation (as well as subsequent generation) we applied the edge grouping 
described at the end of Inference section and then estimate the ${W}_{uv}$ for an edge as the smoothed estimate for the group. 

\subsection{Tasks and experimental results} 

We address three tasks.
(A) {\em Reconstruction accuracy}, (B) {\em Generalization performance} and (C) {\em Discovery and analysis of topic interactions}. 
In task (A), we use the \semi dataset 
to compare the ability of different algorithms to recover the topic and parent of each event and also the network connection strengths. 
In task (B), we use the \real dataset to evaluate 
how well the algorithms fit held-out data after their parameters have been estimated using training data.
In task (C), we investigate the topical interaction matrix generated by \HMHP to find interesting and actionable insights about information cascades in the \real dataset. 
Finally, we also briefly demonstrate the scalability of the inference algorithm for \HMHP.
We next describe the experimental setup for each of the tasks and then the results. 

\subsubsection{Reconstruction Accuracy}
We address three reconstruction tasks: {\em network reconstruction}, {\em parent identification} and {\em topic identification}.
Since the gold-standard values are not known for the \real data, we address these tasks only on the \semi data.
For network reconstruction, we measure distance between estimated and true ${W}_{uv}$ values. 
We report the error in terms of the median Average Percentage Error (APE), 
defined as 
$\sum_{u,v} \frac{|W_{uv} - \hat{W}_{uv}|}{W_{uv}}$, where, $\hat{W}_{uv}$ is the estimated $W_{uv}$ value. 
For comparison with \HTM, we use Total Absolute Error (TAE) (the sum of the absolute differences between the true and estimated $W_{uv}$ values) which was used in the original paper. 
For cascade reconstruction, given a ranked list of predicted parents for each event, we measure accuracy and recall of the parent prediction.
The accuracy metric is calculated as the ratio of number events for which the parent is identified correctly to the total number of events. 
For topic identification, we take a (roughly balanced) sample of event pairs, 
and then measure (using precision/recall/F1) whether the model accurately predicts the event pairs to be from the same or different topics. 


The model that is closest to ours is \HTM \cite{he:icml15}. 
However, the \HTM assigns each event to a topic distribution instead of a single topic. 
Also, the available \HTM code cannot scale to the size of our \semi dataset.
Therefore, we compared our model with \HTM separately.
We first present comparisons with the other baselines on the \semi dataset.
All the presented numbers are averaged over five different semi-synthetic datasets generated independently.


\begin{table}[]
\footnotesize
\centering
\captionsetup{justification=centering}
\begin{tabular}{|m{0.2\linewidth}|c|c|c|}
\hline
 & \HMHP          & \HTL  & \NWL  \\ \hline
Accuracy & 0.581 & 0.362 & 0.370  \\ \hline\hline 
Recall@1  & 0.595 & 0.373 & 0.380   \\ \hline 
Recall@3 & 0.778 & 0.584 & 0.589   \\ \hline 
Recall@5 & 0.838 & 0.674 & 0.678   \\ \hline 
Recall@7 & 0.87 & 0.73 & 0.733    \\ \hline 
\end{tabular}
\caption{Cascade Reconstruction: accuracy and the recall with top-K (1,3,5,7) candidate parents.}
\label{cas_recon}
\end{table}

\begin{table}[]
\small
\centering
\captionsetup{justification=centering}
\begin{tabular}{|l|l|l|l|}
\hline{}
& \HMHP       & \HTL     & \NWL \\ \hline
Mean APE & 0.448 & 0.565 & 0.552  \\ \hline
Median APE & 0.255 & 0.283 & 0.287  \\ \hline
Mean APE ($N_u \ge 100$) & 0.398 & 0.520 & 0.496  \\\hline
Median APE ($N_u \ge 100$) & 0.235 & 0.265 & 0.264 \\\hline
\end{tabular}
\caption{Network reconstruction error (as fraction).}
\label{net_recon}
\end{table}


Table~\ref{cas_recon} presents the accuracy and the recall@k values for the task of predicting parents, and thereby the cascades. 
For all the algorithms, the recall improves significantly by the time the top-3 candidates are considered. For recall@3, \HMHP performs about $32\%$ better than both the other baselines, whereas recall@1, as well as the accuracy of \HMHP are 
at least 57\% better than corresponding number of the baselines.    

For the network reconstruction task, Table~\ref{net_recon} describes the various APE values for models \HMHP, \HTL, and \NWL. Given that in our simulation, event generation was truncated by limiting the number of events, the last level of events contributes to negative bias affecting the APE, since the algorithms are still trying to assign children to these events. We report both the mean and the median errors for this task. The mean error for \HMHP is about $18\%$ lower than other baselines, and the median error is about $10\%$ lower. We also present the results for the set of nodes where the number of generated children ($N_u$) is at least $100$ (this is an easier task). Here also the mean APE for \HMHP is at least $20\%$ lower.


Table~\ref{top_iden} presents the result of the topic modeling task, measured by calculating the precision-recall over event pairs as defined earlier. The results show that \HMHP performs much better than \HTL and at least $5-6\%$ better than \NWL. 

\begin{table}[]
\small
\centering
\captionsetup{justification=centering}
\begin{tabular}{|l|l|l|l|}
\hline
Topic & \HMHP & \HTL  & \NWL  \\ \hline
Precision &	0.893 & 0.123 & 0.781 \\ \hline
Recall & 0.746 & 0.367 & 0.752 \\ \hline
F1    & 0.811 & 0.18 & 0.765 \\ \hline
\end{tabular}
\caption{Event Topic Identification (P/R/F1)}
\label{top_iden}
\end{table}

The above experiment confirms that joint inference of topics, parents, and network strengths by modeling topical interactions is more accurate than that ignoring topical interactions and also  decoupled inference.  While the trend is not surprising as the data was generated using our model, the numbers confirm that when topical interactions patterns are present in the data, our inference algorithm is able to detect and make use of these.
Also, the margin of improvement over the baselines for all tasks is both significant in magnitude as well as statistically significant, showing the importance of topical interactions and joint inference for reconstructions tasks.

\paragraph{Comparison with \HTM} 
\HTM uses a Hawkes process to model the event times, as well as a topic model that takes into account the dependence of child event topics on parent event topics. While there is no notion of topic-topic interaction in \HTM, it is the closest baseline to our model. Therefore we do a more exhaustive comparison against \HTM.
Unfortunately, the available code that we had for \HTM needed approximately half a day for the ArXiv dataset ($\sim$37000 events), and did not scale to our larger datasets. 
So we performed experiments on datasets very similar to the ones used in~\cite{he:icml15}.  
Also, a key difference between \HTM and \HMHP is that, in \HTM each document is a mixture of topics (admixture model), whereas in HMHP each document is generated with a single topic (mixture model). So, we compare \HMHP and \HTM with respect to the quality of network and cascade reconstruction capabilities only and not with respect to topic identification task. To measure the quality of network reconstruction, the evaluation metric used is Total Absolute Error (TAE).

We experimented with synthetic data generated both by the \HMHP and the \HTM models.
We first generate a collection of datasets assuming the \HMHP generative model with two different diffusion networks used in the \HTM paper~\cite{he:icml15}. 
\begin{itemize}
\item \emph{Circular diffusion network:} We use the underlying graph $G=(V,E)$ to be one where $|V| = 10$ and the set of edges are $E = \{(i,j) | ~ j=i ~ or ~ j = (i + 1)\mod |V| \}$. The user-user influence matrix $W_{uv}$ is defined as follows: $W_{ii} = 0.3$ and $W_{ij} = 0.15$, where $j=(i+1)~\mod~10$. Number of topics $k = 10$, with topic $i$ being the only preferred topic of User $i$. Topic word distribution vector for each topic, as well as the topic-topic interaction vectors are drawn from Dirichlet distributions with parameters $0.1$ and $0.01$ respectively. Data is generated for various observation window length ranging from $1000$ to $5000$. For the window length 1000 around 350 events get generated and for window length 5000 around 1900. For each observation window length we generate five different datasets and report the mean TAE (total absolute error) for network reconstruction and mean accuracy for cascade reconstruction. Table \ref{net_recon_HTM_HMHP} shows TAE in network reconstruction for \HTM and \HMHP. The TAE for \HMHP is about half of the TAE of \HTM.  Table \ref{cas_recon_HTM_HMHP} presents the percentage of correctly identified parents for both the models \HTM and \HMHP. Here as well, \HMHP has about $30\%$ better parent identification capability as compared to that of \HTM. 

\item \emph{ArXiv Top-200 Authors Network:} We also perform experiment on the ArXiv citation network dataset. The ArXiv Top-200 authors network dataset is created using the ArXiv high-energy physics theory citation network data from SNAP\footnote{http://snap.stanford.edu/data/cit-HepTh.html (we have obtained our version of dataset from the authors of \HTM \cite{he:icml15})}. We restrict ourselves to a graph formed by the top 200 authors in terms of the number of publications. This graph is similar to the ArXiv Top-200 authors dataset mentioned in \cite{he:icml15}. Topic distributions, topical interactions, and the user-topic preferences are sampled from $Dir(0.1)$, $Dir(0.01)$, and $Dir(0.01)$ respectively. We generate five different datasets each with window length of 25000. The number of events that get generated here are around 37000. Here as well we compare \HTM and \HMHP with respect to the quality of network and cascade reconstruction. Table \ref{net-cas-recon-arxiv-htm-hmhp} presents the mean of (five datasets) TAE and the parent prediction accuracy values for both \HTM and \HMHP models. Here as well, TAE for \HMHP is about $40\%$ lesser than that of \HTM and parent prediction accuracy is at least $30\%$ better. 

\item \emph{Data Generation using \HTM:} For all the above mentioned synthetic experiments, the data was generated according to \HMHP model. In this experiment, we generate data according to \HTM model and compare the performances of \HTM and \HMHP models. Here as well we generate data for window lengths varying from 1000 to 5000, and for each window length five different datasets are generated. The graph or the network used in this experiment is same as that of circular diffusion network mentioned earlier. The document generated for each event is a short document with length sampled from $Poisson(10)$. Table \ref{net-recon-htm-gen-events} shows the average TAE in network reconstruction for both \HTM and \HMHP. Here as well the error in case of \HMHP is half of the error in \HTM. Similarly, Table \ref{cas-recon-htm-gen-events} presents the average parent identification accuracy for both \HTM and \HMHP. The accuracy for \HMHP is at least $14-15\%$ better than that of \HTM.
\end{itemize}

In summary, we see that \HMHP outperforms \HTM quite significantly in parent identification and network strength reconstruction, even when the underlying generative model is that of the \HTM.
We believe there are two reasons for this.
First, a mixture model is a more appropriate model for short documents and can be estimated with greater confidence. 
Secondly, the topic-topic interaction matrix provides crucial additional evidence for parent identification under uncertainty, and this in turn leads to more accurate network strength reconstruction.
In our remaining experiments, we do not evaluate the \HTM further since it does not scale for the datasets that we use.

\begin{table}[]
\centering
\captionsetup{justification=centering}
\begin{tabular}{|l|l|l|l|l|l|}
\hline
Window Length & 1000        & 2000        & 3000        & 4000        & 5000        \\ \hline
\HTM           & 2.811 & 1.982 & 1.464 & 1.292 & 1.351 \\ \hline
\HMHP          & 1.297   & 0.925  & 0.677   & 0.646   & 0.657  \\ \hline
\end{tabular}
\caption{Network Reconstruction Error as TAE (Cycle Graph with 10 nodes)}
\label{net_recon_HTM_HMHP}
\end{table}

\begin{table}[]
\centering
\captionsetup{justification=centering}
\begin{tabular}{|c|c|c|c|c|c|}
\hline
Window Length & 1000        & 2000       & 3000        & 4000        & 5000        \\ \hline
\HTM           & 0.681 & 0.687 & 0.712 & 0.716 & 0.708 \\ \hline
\HMHP          & 0.926   & 0.924   & 0.95   & 0.94   & 0.935   \\ \hline
\end{tabular}
\caption{Cascade Reconstruction Accuracy (Cycle Graph with 10 nodes)}
\label{cas_recon_HTM_HMHP}
\end{table}

\begin{table}[]
\centering
\captionsetup{justification=centering}
\begin{tabular}{|c|c|c|}
\hline
     & Network Reconstruction (TAE) & Cascade Reconstruction \\ \hline
\HTM  & 38.812             & 0.698            \\ \hline
\HMHP & 23.151            & 0.951              \\ \hline
\end{tabular}
\caption{Arxiv Top-200 authors Graph - TAE for Network Reconstruction and Accuracy for Cascade Reconstruction}
\label{net-cas-recon-arxiv-htm-hmhp}
\end{table}

\begin{table}[ht]
\centering
\captionsetup{justification=centering}
\begin{tabular}{|l|l|l|l|l|l|}
\hline
Window Length & 1000        & 2000        & 3000        & 4000        & 5000        \\ \hline
\HTM  & 3.167 & 2.377 & 2.014 & 1.964 & 1.519 \\ \hline
\HMHP & 1.696  & 1.200  & 1.168  & 1.396   & 1.243  \\ \hline
\end{tabular}
\caption{Network Reconstruction TAE for Cycle  (\HTM generated events - short documents)}
\label{net-recon-htm-gen-events}
\end{table}

\begin{table}[ht]
\centering
\captionsetup{justification=centering}
\begin{tabular}{|l|l|l|l|l|l|}
\hline
Window Length & 1000        & 2000        & 3000      & 4000        & 5000       \\ \hline
\HTM & 0.575 & 0.588 & 0.61 & 0.618 & 0.628 \\ \hline
\HMHP & 0.716   & 0.730   & 0.736 & 0.730   & 0.748  \\ \hline
\end{tabular}
\caption{Cascade Reconstruction Accuracy for Cycle (\HTM generated events - short documents)}
\label{cas-recon-htm-gen-events}
\end{table}

\subsubsection{(B) Data fitting quality} 
We now evaluate goodness of fit for different models in the held out setting. 
All the models were trained on 500K events and held-out log-likelihood was calculated on the 500K events immediately following the training events. 
Since this does not require gold-standard labels, this can be evaluated on \real as well as \semi data.
Here we present results only for the \real data, since that is the harder task for our model. 
Table-\ref{holl} shows the values of the log-likelihood for all the three models. 
We observe that the likelihood number for \HMHP is roughly 5\% better than the ones for both the 
\HTL and \NWL. 
We also observe the test likelihood of \HMHP improves with the number of topics.



\begin{table}[]
\centering
\begin{tabular}{|c|l|l|l|l|}
\hline
\multicolumn{1}{|l|}{\#Topics} & Log-Likelihood & \HMHP         & \HTL         & \NWL     \\ \hline
\multirow{3}{*}{25} & Content & -30499278 & -33356945  & -30532938  \\ \cline{2-5} 
& Event Time & -4236958 & -4042903  & -4299630   \\ \cline{2-5} 
& Total & -34736237 & -37399849  & -34832568 \\ \hline \hline
\multirow{3}{*}{50} & Content & -30141081 & -33427354  & -30089733  \\ \cline{2-5} 
&Event  Time & -4288438 & -4510072  & -4343571  \\ \cline{2-5} 
& Total & -34429519 & -37937426 & -34433305   \\ \hline \hline
\multirow{3}{*}{75}& Content & -29860909 & -33433922 & -29861050  \\ \cline{2-5} 
& Event Time & -4285293 & -4510535  & -4373736  \\ \cline{2-5} 
& Total & -34146202 & -37944457 & -34234787  \\ \hline
\end{tabular}
\caption{Held-out Log-Likelihood, shown individually for tweet content and activation times, as well as in aggregate.}
\label{holl}
\end{table}

\begin{table}[]
\centering
\begin{tabular}{|c|c|c|c|c|}
\hline
 & \multicolumn{2}{c|}{\textbf{100K Events}} & \multicolumn{2}{c|}{\textbf{500K Events}} \\ \hline
\#Topics & Topic Time & Total Time & Topic Time & Total Time \\ \hline
25 & 3 & 6 & 18 & 44 \\ \hline
50 & 6 & 9 & 30 & 56 \\ \hline
75 & 8 & 11 & 41 & 67 \\ \hline
100 & 12 & 15 & 56 & 82 \\ \hline
150 & 16 & 19 & 81 & 107 \\ \hline
\end{tabular}
\caption{Time per iteration (in secs) taken by \HMHP on 100K and 500K events. Parent update time for 100K events is $\sim$3 secs and for 500K events is $\sim$25 secs, User-User influence update time is $\sim$0.09 secs and $\sim$1 sec for 100K and 500K events respectively.}
\label{hmhp_scalability}
\end{table}

Just as for reconstruction accuracy on \semi data, we see that modeling and detecting topical interaction patterns jointly with other tasks leads to significantly better generalization performance for real data. 
From this we may claim (of course without direct validation) that \HMHP would perform more accurate reconstruction as well on \real data. 

\subsubsection{(C) Discovery and Analysis of topic interactions.} 
The final task is to discover statistically significant topical interactions from textual information cascades and investigate what actionable insights can be drawn from such topical patterns. 
We stress that this task can only be performed using \HMHP, and as such there is no baseline algorithm for this task.
The first set of results on the \semi data demonstrates that when the data has realistic parameters, but is generated from the model that we hypothesize, \HMHP outperforms the baselines. 
The generalization experiments further confirm that even for real data \HMHP performs better than the baselines, suggesting that real data indeed better matches our modeling assumptions about interacting topics.
This lends credibility to the topical interactions discovered by \HMHP model, even though there is no ground truth for these.

In order to demonstrate the usefulness of the topic-topic interaction matrix, we perform three types of analysis using this matrix. We first select a set of topic-topic relationships as anecdotes. 
Next we demonstrate how we can get insights on topic drift in cascades using a hubs and authorities analysis on the topic-topic interaction matrix. 
Finally, we use a personalized pagerank based analysis to discover related topics for any given topic and discuss how this can lead to new strategies for user targeting and influence maximization problems for spreading conversations about any topic. 

\begin{itemize}

\item{\textbf{Anecdotal parent-child topics:}} To identify interesting examples of topic-topic interaction, we first ensure that the topic pairs do not correspond to the same underlying topic that our model inappropriately partitioned into multiple finer topics. Such a phenomenon would cause a block-diagonal structure in the topic-topic interaction matrix. We first identify the most asymmetric topic pairs by sorting using $\mc{T}_{kk'} - \mc{T}_{k'k}$. 
Such pairs cannot conceivably come from topic splitting. 
To illustrate the topics, we find their top five hashtags. 
Table~\ref{anec2} shows examples of such topic pairs. 
In the first row, the parent and child hashtags are related to American football and baseball. These are different topics, but indicate that tweets related to a particular sport (football) trigger tweets about some other sport (baseball). 
Alternatively, there can be users who participate in the discussions related to both the sports.  
Table~\ref{anec1} shows a few actual parent-child tweets from the selected asymmetric topic pairs. It is important to stress that these tweet pairs are not retweets. Also, though all these tweet pairs clearly on related topics, often (like the first pair about {\em MH370}) these do not share any hashtags or even other significant words, which some naive strategy might use to detect such relationships. Therefore, in both the cases, we see evidence of interesting topic interactions that are discovered by our model. 

Such significant topical interactions can then serve as input to various further tasks e.g. getting a global view of ongoing discussions on some events, possibly better estimations of which topics are going viral etc. Indeed, we can argue that topics, rather than hashtags, represent the correct granularity for such analysis.

\item{\textbf{Hub and authority topics:}} We next perform a hub-topic analysis of the topic-topic interaction matrix by representing it as a bipartite graph-- one side being the `hubs' and the other side the `authorities'.
Note that a topic can potentially have both a hub and an authority representation. To eliminate some of the estimation noise, we consider only the edges with $\mc{T}_{kk'} > 0.1$ and also ignore the dominating diagonal terms i.e. $\mc{T}_{kk}$. Next, we run the HITS algorithm~\cite{kleinberg1999authoritative} to find out the prominent hubs and authorities. Given that most of the cascades are short, we can think of the hubs and authorities analysis as a ``one-step approximation" of the topical Markov chain.  
We found that the hub scores were more or less uniformly distributed whereas the authority scores were more skewed-- showing that while cascades can start with any specific topic, they tend to converge to the generic ones.
This intuition is strengthened by Table \ref{anec_hubs_auths} which shows $5$ topics with the highest hub scores and $5$ topics with the highest authority scores. 
We note that the hub topics in general appear more focused than the authority topics. 
One way of explaining is that the authority topics here are more representative of final or terminal topics of the cascades, while the hubs capture starting or initial topics in cascades. Table~\ref{anec_ppr} then has an intuitive explanation that conversations start with specific topics but then often diverge into more general ones. 

\item{\textbf{Modeling Topic Drifts via random walks:}} One of the prominent applications of information cascade modeling is in targeting-- how to intelligently select a small set of triggers in the network that can encourage conversations/tweets about topics or hashtags that the advertiser is interested about. 
We claim that our model, by capturing the topic-topic interactions, provides alternate topics or hashtags for such advertisers to exploit. 
To demonstrate this, we choose few of the topics and then run personalized pagerank~\cite{page1998pagerank} from these topic nodes on the Markov chain underlying the topic-topic interaction matrix. 
Table \ref{anec_ppr} shows the results of running personalized pagerank from example start topics. 
For each topic in the first column of the Table \ref{anec_ppr}, the second column contains the $3$ topics (other than the topic itself) with highest personalized pagerank. For the examples in the first $3$ rows, most of the top $3$ transitioned topics are the high authority score topics (more general ones). However, after removing some of these topics from the transition matrix, the top $3$ transitioned topics (right column last 3 rows) have topics that are directly related to the start topic. For e.g. the start topic in the fourth row is about Russia, Syria, and Ukraine, and the transitioned topics have hashtags that are related to politics in US. This gives an hint about the connection between US, Russia, Syria, and Ukraine in Twitter conversations. 
Similarly, in the last row the start topic is about TV shows, and the transitioned topics also contain hashtags which are about some other TV shows. 

One interpretation of this is that the average conversation that starts with any topic from the first column, can be expected to drift to one of the topics in the corresponding row. 
While this knowledge can have many applications, it is particularly useful for advertisers who are interested in promoting specific keywords or hashtags. 
For instance, if the advertiser is interested in incentivizing conversations about a target topic in the second column, then apart from the users whose interest matches this specific topic, it is now also possible to try to encourage and incentivize users whose user-topic interest matches the corresponding start topic, since such conversations drift to the desired target topic with high probability.

\end{itemize}

\begin{table}
\footnotesize
 \centering
 \captionsetup{justification=centering}
 \begin{tabular}{|p{0.35\linewidth}|p{0.45\linewidth}|}
 \hline
 \textbf{Parent topic hashtags} & \textbf{Child topic hashtags}\\
 \hline
steelers, browns, seahawks, fantasyfootball, nfl & mlb, orioles, rays, usmnt, redsox\\
\hline
renewui, ableg, aca, stopcombatingme, obamacare & 
uniteblue, tx2014, saysomethingliberalin4words, tcot, gophatesvets\\
\hline
idf, rnb, bds, gaza, israel & russia,iran,syria,crimea,ukraine\\
\hline
thewalkingdead, theamericans, onceuponatime, houseofcards, tvtag & arrow, agentsofshield, truedetective, longislandmedium, tvtag\\
\hline
mlb, packers, fantasyfootball, nfl, redsox & 
tblightning, sabres, canucks, nyr, iahsbkb\\
\hline
egypt,gaza,israel,syria,iran & 
kiev, putin, russia, crimea, ukraine\\
\hline
 \end{tabular}
 \caption{Sample hashtags from asymmetric topics pairs.}
\label{anec2}
\end{table}

\begin{table}[t]
\footnotesize
\centering
\captionsetup{justification=centering}
 \begin{tabular}{|p{0.45\linewidth}|p{0.45\linewidth}|}
 \hline
 \textbf{Parent Tweet} & \textbf{Child Tweet} \\
 \hline
 [\#MASalert] Statement By Our Group CEO, Ahmad Jauhari Yahya on MH370 Incident. Released at 9.05am/8 Mar 2014 MY LT & 
 Missing \#MalaysiaAirlines flight carrying 227 passengers (including 2 infants) of 13 nationalities and 12 crew members.\\
 \hline 
 Investigators pursuing notion the 777 was diverted ``with the intention of using it later for another purpose.'' & If \#MH370 did fly for an additional 4 hrs as reported by WSJ, it could be anywhere in this circle. \\ 
 \hline
 Wanna be a part of GOTV in the first special election of year? Hop on our caller for Alex Sink in Florida: \#FL13 & Why \#CPAC Isn’t As Grassroots As You Think: \#independents \#moderates \#tcot \#gop \#ccot \#tlot \#pjnet \#uniteblue \#p2 \\
 \hline
 The academy is using Sara Jones in memoriam to troll their website. \#oscars & What is going on with Travolta's center part? \#oscars2014 \\
 \hline
 Certain people are ruining their reputations tonight-really sad! \#Oscars & I should host the \#Oscars just to shake things up - this is not good! \\
 \hline
 mt@conor64 Obama Is Complicit in Covering Up the Truth About \#CIA Torture \#p2 \#tcot \#topprogs \#teaparty \#uniteblue & The moral of `Green Eggs and Ham' is lost on Ted Cruz and Sarah Palin. \#UniteBlue and \#TryNewThings \\ 
 \hline
 Gellman:My definition of whistle blowing:are you shedding light on crucial decision that society should be making for itself. \#snowden & Gellman we are living inside a one way mirror,they \& big corporations know more and more about us and we know less about them \#sxsw \\
 \hline
 \end{tabular}
 \caption{Example Parent-Child Tweets}
\label{anec1}
 \end{table}

\begin{table}
\footnotesize
 \centering
 \captionsetup{justification=centering}
 \begin{tabular}{|p{0.43\linewidth}|p{0.47\linewidth}|}
 \hline
 \textbf{Hubs (Top-5)} & \textbf{Authorities (Top-5)}\\
 \hline
blogtalkradio, escort, buckeyes, ohiostate, ep2014 & liestoldbyfemales, ifeverybodyran, teenmom2, tni, fail\\
\hline
photo,craftbeer,beer,sxswi,yelp & sxswi,expowest,dx32014,sxsw,sxsw14\\
\hline
sxswedu, ncties14, edtech, dml2014, edchat & tbt, throwbackthursday,  wcw, throwback, 100happydays\\
\hline
rhlaw, clinton, r4today, ecommerce, cadem14 & agentsofshield, arrow, tvtag, supernatural, chicagoland\\
\hline
pinit, runtastic, windowsazure, iphone, qconlondon & marketing, socialmedia, seo, contentmarketing, smmw14\\
\hline
 \end{tabular}
 \caption{Hashtags from Hub and Authority Topics}
\label{anec_hubs_auths}
\end{table}

\begin{table}
\footnotesize
 \centering
 \captionsetup{justification=centering}
 \begin{tabular}{|p{0.23\linewidth}|p{0.57\linewidth}|}
 \hline
 \textbf{Source Topic hashtags} & \textbf{Hashtags from top-3 transitioned topics}\\
 \hline
[ukraine, crimea, russia, putin, syria] & [liestoldbyfemales, ifeverybodyran, teenmom2, tni, fail], [sxswi, expowest, dx32014, sxsw, sxsw14], [marketing, socialmedia, seo, contentmarketing, smmw14]
 \\ \hline
[rhlaw, clinton, r4today, ecommerce, cadem14] & [sxswi, expowest, dx32014, sxsw, sxsw14], [irs, tcot, teaparty, gophatesvets, uniteblue], [liestoldbbyfemales, ifeverybodyran, teenmom2, tni, fail]
\\ \hline
[agentsofshield, arrow, tvtag, supernatural, chicagoland] & [liestoldbyfemales, ifeverybodyran, teenmom2, tni, fail],  [idol, bbcan2, havesandhavenots, pll, thegamebet], [sxswi, expowest, dx32014, sxsw, sxsw14]
\\ \hline \hline
\multicolumn{2}{|c|}{\textbf{After removing 7 generic topics}}
\\ \hline \hline
[ukraine, crimea, russia, putin, syria] & [utpol, raisethewage, cdnpoli, obamacare, aca], [irs, tcot, teaparty, gophatesvets, uniteblue], [worldbookday, amwriting, books, litfestlive]
\\ \hline
rhlaw, clinton, r4today, ecommerce, cadem14] & [irs, tcot, teaparty, gophatesvets, uniteblue], 
[soundcloud, hiphop, mastermind, nowplaying, music], 
[iahsbkb, nba, iubb, lakers, rockets]
\\ \hline
[agentsofshield, arrow, tvtag, supernatural, chicagoland] & [idol, bbcan2, havesandhavenots, pll, thegamebet], [tvtag, houseofcards, agentsofshield, arrow, theamericans], [soundcloud, hiphop, mastermind, nowplaying, music], 
\\ \hline
\end{tabular}
\caption{Example related topics using personalized pagerank.}
\label{anec_ppr}
\end{table}

\paragraph{Runtime of \HMHP} Finally, we comment about the execution times of our algorithm.
The \HMHP inference algorithm were coded using C++ were run on a 12 core Xeon E5 machine with 32GB of memory, with no special optimizations other than the compiler optimization of level $O2$. 
Table~\ref{hmhp_scalability} shows that even without any special optimization our algorithm is quite scalable owing to the efficient collapsed Gibbs Sampling strategy, and suggests that it can be used for web-scale analysis. 

To summarize, using multiple semi-synthetic datasets we show that the \HMHP significantly outperforms state-of-the-art baselines including the \HTM in cascade and network reconstruction by virtue of modeling topical interactions and joint inference of parents, topics and network strengths. 
It also generalizes significantly better for real Twitter data, demonstrating that topical interactions is indeed a better model for real cascade of tweets. The scalability of \HMHP makes it amenable to real web-scale applications. Finally, we  analyzed the topical interactions identified by \HMHP to provide guidelines about how this could be used for influence maximization, as well as unearthing new insights about topic drift in cascades.

%% file: rw.tex
There has been a lot of work on network reconstruction based on the observations of event times~\cite{rodriguez14uncovering,wang14mmrate}--- these models often ignore the content information of the events. Our work shows that such content information, when present, can profitably be used for a better estimation of the network strengths.
~\cite{barbieri2017survival} studies a related model in which activation times and some side information (e.g. tweet content) about cascades are observed, but not the set of users.  

The Dirichlet Hawkes Process (DHP) \cite{du:kdd15} and the Hawkes Topic Model (HTM) \cite{he:icml15} both extend Hawkes Processes to model textual content associated with events. 
However, neither captures topical interactions.
Like our model, the DHP is a mixture model in that it assigns to each event a single topic.
However, it lacks the notion of a unique parent for any event --- instead past topics are reused randomly based on their recency.
It also lacks any notion of users or networks.
Our model is most similar to the HTM.
But while the HTM forces parent and child events to be topically close, it does not capture any parent-child topical patterns beyond this.
One consequence is that no two events are topically identical, while we can group events according to assigned topics and, additionally, parent-child relations according to topic pairs.
HTM's document model is an admixture of topics, which is more powerful in general, but in the context of short documents such as tweets, this complexity is not as necessary, as we demonstrate in our experiments.

The Correlated Topic Model (CTM) \cite{lafferty:nips05} captures topic correlations within individual documents, unlike our modeling of topical transition patterns in event sequences. 

Models for sequence data have blended Markovian dynamics with topic models \cite{griffiths:nips05,gruber:aistats07,Du2012,barbieri13prob}.
This thread, often called sequential LDA / topic models, are focused on modeling richer sequential structure within a single document, either by assuming segments being generated from individual topics, or a sequential sampling of word sequences. 
However, this thread has not looked at cascades of events with associated time stamps and networked users.
Our contribution is in merging this line of research with the modeling of information cascades.


%% file: conclusion.tex
In summary, we propose a generative model for information diffusion cascades accounting for topical interactions, by coupling a Network Hawkes process with a Markov Chain over topics for diffusion paths. 
This enables us to fit real Twitter conversations better, and the use of topic interactions and collective inference using our model also leads to more accurate reconstruction of network strengths, diffusion paths and event topics.
Using comparisons on a number of datasets, we show that our model outperforms the existing baselines on the standard metrics. 
On top of this, using our model, we are able to derive insights about topical interactions, which existing models cannot. 
This can potentially lead to new actionable strategies for user targeting and influence maximization.